\documentclass[10pt,twocolumn,letterpaper]{article}

\usepackage{cvpr}              % To produce the CAMERA-READY version
\usepackage{booktabs}   % 用于生成专业美观的表格线
\usepackage{multirow}   % 用于创建跨越多行的单元格
\usepackage{makecell}   % 用于在单元格内换行和更精细的对齐控制
\usepackage{graphicx}   % \resizebox 需要此宏包
\usepackage{booktabs} % 用于创建更美观的表格线
\usepackage{threeparttable} % 核心宏包

\definecolor{cvprblue}{rgb}{0.21,0.49,0.74}
\usepackage[pagebackref,breaklinks,colorlinks,allcolors=cvprblue]{hyperref}

\def\paperID{} % *** Enter the Paper ID here
\def\confName{CVPR}
\def\confYear{2026}

\title{Learning to Focus and Precise Cropping: A Reinforcement Learning Framework with Information Gaps and Grounding Loss for MLLMs}

\author{Xuanpu Zhao\textsuperscript{\rm 1,2}\quad Zhentao Tan\textsuperscript{\rm 3}\quad Dianmo Sheng\textsuperscript{\rm 1,2}\quad Tianxiang Chen\textsuperscript{\rm 1,2}\quad Yao Liu\textsuperscript{\rm 3}\\Yue Wu\textsuperscript{\rm 3}\quad Tao Gong\textsuperscript{\rm 1,2}\thanks{Corresponding author.}\quad Qi Chu\textsuperscript{\rm 1,2}\quad Nenghai Yu\textsuperscript{\rm 1,2}  \\
\textsuperscript{\rm 1}School of Cyber Science and Technology, University of Science and Technology of China\\
\textsuperscript{\rm 2}Anhui Province Key Laboratory of Digital Security\quad\textsuperscript{\rm 3}Independent Researcher\\
{\tt\small zhaoxuanpu@mail.ustc.edu.cn}\quad{{\tt\small zhentaotan5@gmail.com}}\quad{\tt\small tgong@ustc.edu.cn}
}

\begin{document}
\maketitle 
\begin{abstract}
To enhance the perception and reasoning capabilities of multimodal large language models in complex visual scenes, recent research has introduced agent-based workflows. In these works, MLLMs autonomously utilize image cropping tool to analyze regions of interest for question answering. While existing training strategies, such as those employing supervised fine-tuning and reinforcement learning, have made significant progress, our empirical analysis reveals a key limitation. We demonstrate the model's strong reliance on global input and its weak dependence on the details within the cropped region. To address this issue, we propose a novel two-stage reinforcement learning framework that does not require trajectory supervision. In the first stage, we introduce the ``Information Gap" mechanism by adjusting the granularity of the global image. This mechanism trains the model to answer questions by focusing on cropped key regions, driven by the information gain these regions provide. The second stage further enhances cropping precision by incorporating a grounding loss, using a small number of bounding box annotations. Experiments show that our method significantly enhances the model's attention to cropped regions, enabling it to achieve state-of-the-art performance on high-resolution visual question-answering benchmarks. Our method provides a more efficient approach for perceiving and reasoning fine-grained details in MLLMs. Code is available at: \url{https://github.com/XuanPu-Z/LFPC}.
\end{abstract}    
\section{Introduction}
\label{sec:intro}

\begin{figure*}[t]
\centering
\includegraphics[width=\linewidth]{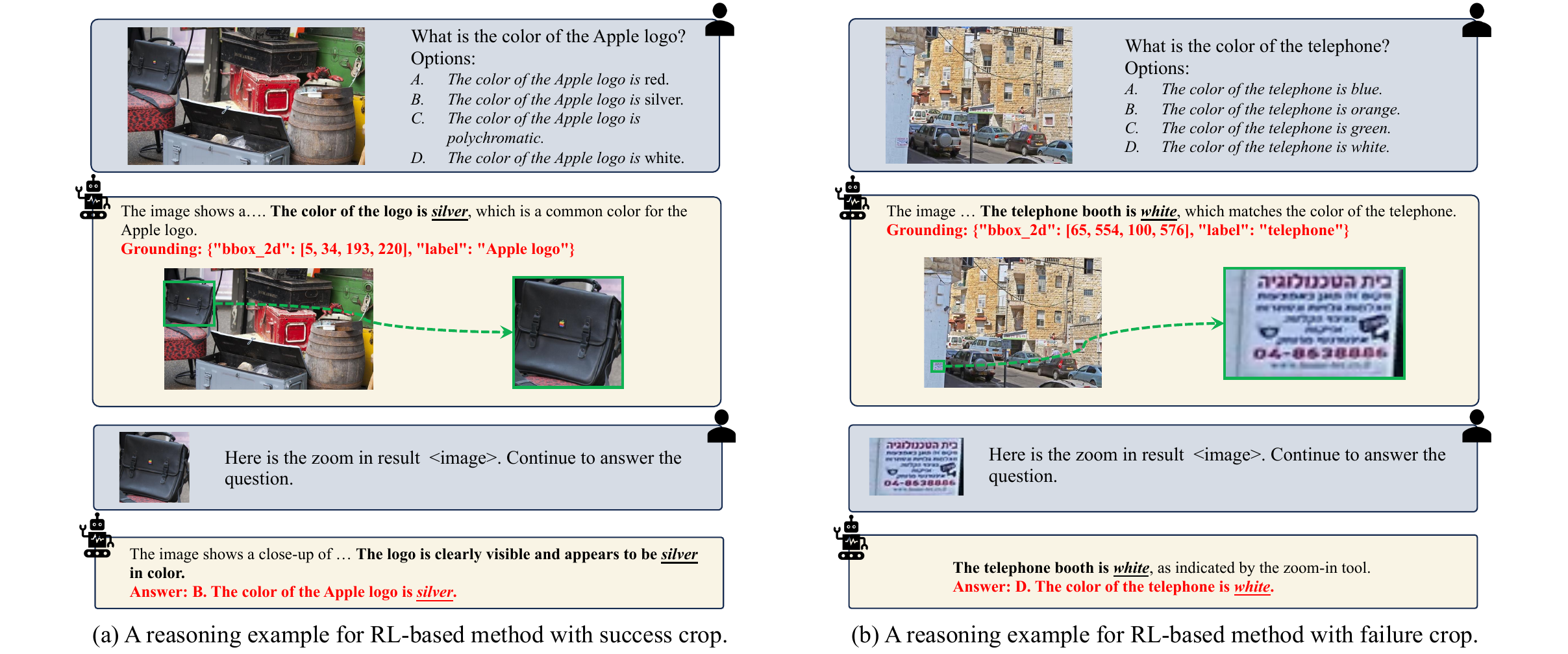} 
\caption{Reasoning example for RL-based methods. In example (a), the model crops the region correctly but still fails to answer the color of the logo. While in example (b), the model confuses the sign with a telephone booth even after zooming in.} 
\label{fig:intro}
\end{figure*}

Multimodal Large Language Models (MLLMs) have demonstrated remarkable capabilities in understanding and reasoning about the visual world, achieving impressive performance across a wide range of vision-language tasks \cite{hudson2019gqa, singh2019towards, liu2024mmbench, fu2024blink}. However, their ability to perform fine-grained perception and reasoning in complex visual scenes remains a significant challenge. Especially when the object of reasoning is small or obscured by a complex background \cite{wang2025divide, wu2024v}, the model struggles to accurately locate the target and attend to detailed cues in a single inference.
%This limitation often stems from the fixed, and often low, resolution of input images, which forces the model to process a condensed representation of the entire scene, inevitably losing critical local information.

To address these limitations, a promising direction in recent research has been the introduction of agent-based workflows \cite{wang2025pixel, zhang2025chain, lai2025mini,zheng2025deepeyes}. In this paradigm, the MLLM is not merely a passive observer but an active agent that can use external tools to dynamically explore visual input. A particularly effective approach involves empowering the MLLM with a ``cropping tool", allowing it to autonomously identify and zoom in on specific regions of interest to gather detailed information necessary for answering complex questions. These agentic MLLMs have achieved substantial progress through two primary training strategies: a hybrid of Supervised Fine-Tuning (SFT) and Reinforcement Learning (RL) \cite{wang2025pixel, zhang2025chain, lai2025mini},  or a pure Reinforcement Learning approach \cite{zheng2025deepeyes}.

Despite their success, existing training methodologies have some inherent limitations. The SFT+RL hybrid approach heavily relies on generating vast quantities of reasoning trajectories from a powerful, proprietary teacher MLLM. This process is not only computationally expensive and time-consuming but also creates a performance ceiling, as the student model's capabilities are inherently upper-bounded by those of its teacher. While pure RL methods circumvent the need for a teacher model, they introduce a subtler yet critical issue that we observed, which is also present to some extent in hybrid approaches. Taking DeepEyes \cite{zheng2025deepeyes}, a typical RL-based model, as a sample, we discovered a recurring pattern in its reasoning process: the model often predicts the final answer before executing a crop, and subsequently uses the cropped region merely to confirm its pre-existing conclusion, as shown in Figure ~\ref{fig:intro}.
We hypothesize that this ``answer first, crop later" behavior indicates that the model is often performing a perfunctory tool call for cropping, rather than genuinely leveraging the information within the cropped region to aid its reasoning and question-answering process. To validate this hypothesis, we constructed a dedicated evaluation benchmark, which confirmed that the model's performance is not significantly dependent on the content of the cropped region.

To tackle this fundamental challenge of ensuring authentic reliance on tool-based observations, we propose a novel two-stage pure reinforcement learning framework. Our core objective is to compel the model to actively seek out and utilize information from cropped regions. In the first stage, we introduce an ``Information Gap Mechanism". In contrast to previous work that directly feeds original high-resolution images into the model, we strategically downsample the initial input images. The degree of downsampling is determined by the model's own uncertainty; we select an appropriately low resolution such that the model yield answers inconsistent with those obtained at higher resolutions. However, when the agent decides to use its cropping tool, the region is extracted from the original, full-resolution image. This creates a crucial ``information gap" between the low-detail global view and the high-detail local view, making the information within the crop indispensable for a correct answer and thus enhancing the model's focus on the cropped region. In the second stage, to further refine the agent's behavior, we annotate a small number of bounding boxes and introduce a grounding reward signal. This encourages the model to not only use the cropping tool but also to place the cropped region at a more precise location.

Our experiments demonstrate the effectiveness of our method. It significantly improves the model's attention to and reliance on cropped regions, while mitigating the formalistic cropping tool invocation issue observed in previous work. Remarkably, our method achieves state-of-the-art performance under both visual token budgets of 16,384 and 1,024 for the input image. Furthermore, when the visual token budget is limited to 1,024, our method still outperforms other approaches on some benchmarks even when they are allowed 16,384 visual tokens, highlighting both its effective utilization of fine-grained details within cropped regions and its computational efficiency.

\section{Related Work}
\label{sec:related}

\textbf{Multi-modal Large Language Models.} Multimodal large language models (MLLMs) has progressed from early systems that aligned a vision encoder with a large language model \cite{chung2024scaling, touvron2023llama} to more tightly integrated structures enabled by joint training. BLIP-2 \cite{li2023blip} and LLaVA \cite{liu2023visual, liu2024improved} pioneered the alignment of visual features with LLMs using lightweight Q-Former or projectors, while approaches such as LLaVA-OneVision \cite{li2024llava} enhanced visual fidelity by accommodating flexible image resolutions. These advancements have fostered a rich ecosystem of powerful open-source models, including the Qwen-VL \cite{bai2023qwen,wang2024qwen2,bai2025qwen25vltechnicalreport}, and InternVL \cite{chen2024internvl,chen2024far,chen2024expanding,zhu2025internvl3} series. With the maturation of these models, attention has increasingly turned to their reasoning abilities, particularly through Multimodal Chain-of-Thought (MCoT). Research in MCoT reasoning can be broadly categorized into two main paradigms. One paradigm depends on manually designed workflows to structure reasoning \cite{liu2024chain, mondal2024kam}, typically targeting problems like localization \cite{fu2025refocus, wei2025perception} or knowledge retrieval \cite{sun2024visual,li2025imagine}. The other explores RL-based approaches \cite{peng2025lmm, shen2025vlm} to adapt text-centric reasoning for multimodal challenges like spatial reasoning \cite{ouyang2025spacer,li2025star,wu2025reinforcing} and reasoning segmentation \cite{liu2025seg}. 

\noindent \textbf{High-Resolution Visual Question Answering.}
Multimodal Large Language Models (MLLMs) performs poorly on High-Resolution Visual Question Answering \cite{wu2024v,wang2025divide} task which requires them to perceive fine-grained details in complex scenes. To address this, one line of works uses attention maps to analyze import regions \cite{zhang2025mllms,mao2025through}. Another approach represents the image as a hierarchical tree of regions and employs a search algorithm to locate the import regions \cite{li2025dyfo,mao2025through,wang2025pixel}.
Despite demonstrating significant performance improvements, these methods often involve complex pipelines and suffer from low inference efficiency.

\noindent \textbf{Agent-based Methods.}
Recently, a line of work has emerged to address complex visual reasoning by introducing agentic workflows \cite{yao2022react}. These methods can actively invoke an cropping tool, to focus on critical regions and facilitate their reasoning process. These approaches can be broadly categorized into two streams. The first stream employs a two-stage training paradigm: Supervised Fine-Tuning (SFT) followed by Reinforcement Learning (RL) \cite{zhang2025chain, wang2025pixel, lai2025mini}. Initially, a powerful teacher MLLM (e.g., GPT-4o \cite{hurst2024gpt}) generates reasoning trajectories, which include both textual rationales and coordinates of corresponding cropped regions. These trajectories are then used to supervise the fine-tuning of the agent model, which is subsequently optimized further using RL. The second stream relies solely on RL for training \cite{zheng2025deepeyes}, using a dataset that only consists of image-question pairs. However, the SFT+RL paradigm is constrained by the need to generate high-quality trajectories, a process that is both costly and fundamentally capped by the teacher model's capabilities. As for the pure RL methods, we observe that their impressive performance still heavily relies on the original, full input image, with weak attention paid to the cropped regions.

\section{Preliminary Analysis}
\label{sec:formatting}

\begin{figure}[t]
\centering
\includegraphics[width=\linewidth]{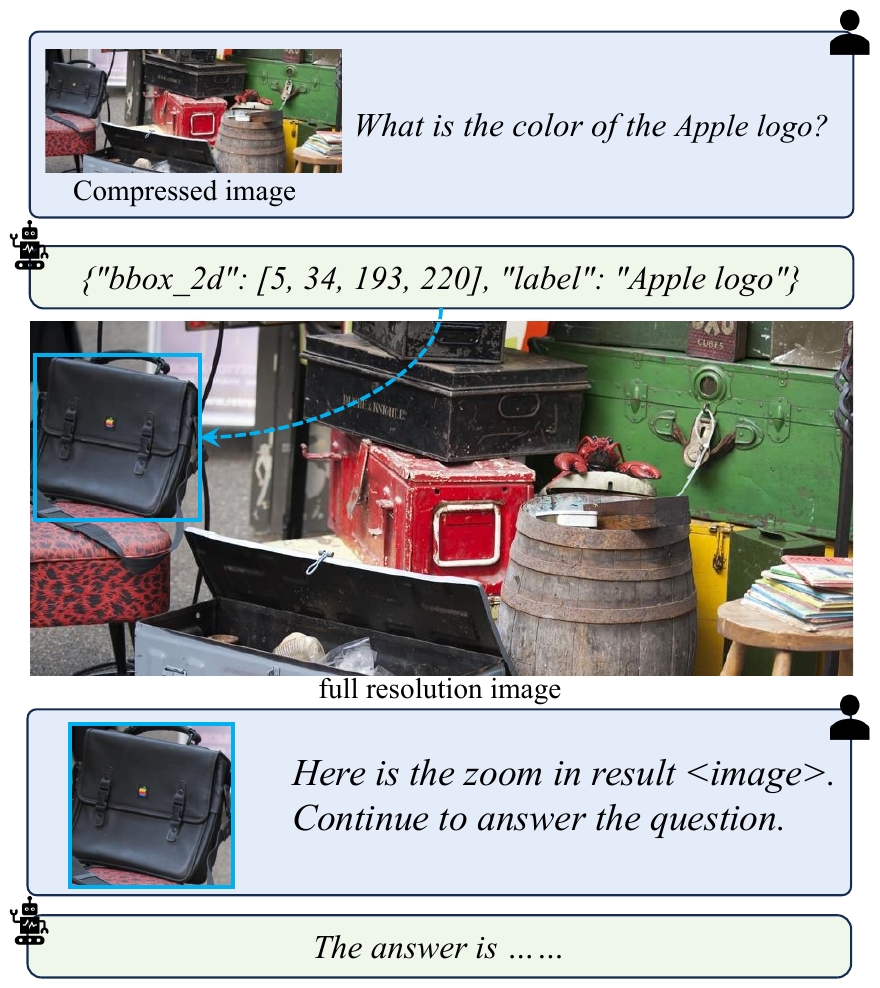} 
\caption{Testing pipeline of agentic-based MLLMs.}
\label{bench}
\end{figure}

\begin{table*}[t] % [H] 强制表格出现在此处; 您也可以用 [htbp] 让 LaTeX 自动选择位置
    \centering % 将整个表格（包括子表）居中

    \begin{subtable}{0.9\textwidth}
    \centering
    % \small % 如果表格过宽，可以取消注释此行以缩小字体
    \begin{tabular}{l ccc | ccc | ccc}
    \toprule
    % 使用 multirow 使 "Model" 单元格跨越两行，并垂直居中。
    % \textbf{} 用于加粗文本。
    % \multicolumn 用于合并列，创建顶层表头。
    \multirow{2}{*}{\textbf{Model \& Setting}} & \multicolumn{3}{c}{HR-Bench 8K (Lite)} & \multicolumn{3}{c}{HR-Bench 4K (Lite)} & \multicolumn{3}{c}{\textbf{$V^{*}$}} \\
    % \cmidrule 从 booktabs 宏包而来，用于绘制跨越指定列的部分水平线，比 \cline 更美观。
    % (lr) 选项使线条左右两端有轻微的缩进，视觉效果更好。
    \cmidrule(lr){2-4} \cmidrule(lr){5-7} \cmidrule(lr){8-10}
    % 第二行表头
    & \textbf{FSP} & \textbf{FCP} & \textbf{Overall} & \textbf{FSP} & \textbf{FCP} & \textbf{Overall} & \textbf{Attr} & \textbf{Spatial} & \textbf{Overall} \\
    \midrule
    % 表格数据行
    DeepEyes prediction                 & 83.0 & 53.0 & 68.0 & 92.0 & 56.0 & 74.0 & 84.3 & 88.2 & 85.9 \\
    DeepEyes Random Noise               & 83.0 & 54.0 & 68.5 & 94.0 & 57.0 & 75.5 & 83.5 & 88.2 & 85.4 \\
    DeepEyes Ground Truth               & 86.0 & 53.0 & 69.5 & 93.0 & 55.0 & 74.0 & 86.1 & 89.5 & 87.5 \\
    \midrule
    CoF-sft prediction                  & 87.0 & 53.0 & 70.0 & 94.0 & 52.0 & 73.0 & 90.4 & 85.5 & 88.5 \\
    CoF-sft Random Noise                & 85.0 & 52.0 & 68.5 & 90.0 & 53.0 & 71.5 & 80.9 & 84.2 & 82.2 \\
    CoF-sft Ground Truth                & 89.0 & 51.0 & 70.0 & 96.0 & 54.0 & 75.0 & 91.3 & 85.5 & 89.0 \\
    \bottomrule
    \end{tabular}
    \caption{Maximum visual tokens limited at 16,384.}
    \label{tab:my_awesome_results_16384}
    \end{subtable}

    \begin{subtable}{0.9\textwidth}
    \centering
    % \small % 如果表格过宽，可以取消注释此行以缩小字体
    \begin{tabular}{l ccc | ccc | ccc}
    \toprule
    % 使用 multirow 使 "Model" 单元格跨越两行，并垂直居中。
    % \textbf{} 用于加粗文本。
    % \multicolumn 用于合并列，创建顶层表头。
    \multirow{2}{*}{\textbf{Model \& Setting}} & \multicolumn{3}{c}{HR-Bench 8K (Lite)} & \multicolumn{3}{c}{HR-Bench 4K (Lite)} & \multicolumn{3}{c}{$V^{*}$} \\
    % \cmidrule 从 booktabs 宏包而来，用于绘制跨越指定列的部分水平线，比 \cline 更美观。
    % (lr) 选项使线条左右两端有轻微的缩进，视觉效果更好。
    \cmidrule(lr){2-4} \cmidrule(lr){5-7} \cmidrule(lr){8-10}
    % 第二行表头
    & \textbf{FSP} & \textbf{FCP} & \textbf{Overall} & \textbf{FSP} & \textbf{FCP} & \textbf{Overall} & \textbf{Attr} & \textbf{Spatial} & \textbf{Overall} \\
    \midrule
    % 表格数据行
    DeepEyes prediction    & 68.0 & 59.0 & 63.5 & 78.0 & 54.0 & 66.0 & 75.7 & 76.3 & 75.9 \\
    DeepEyes Random Noise & 62.0 & 57.0 & 59.5 & 73.0 & 54.0 & 63.5 & 73.0 & 77.6 & 74.8 \\
    DeepEyes Ground Truth & 70.0 & 59.0 & 64.5 & 79.0 & 54.0 & 66.5 & 77.4 & 77.6 & 77.5 \\
    \midrule
    CoF-sft prediction    & 68.0 & 53.0 & 60.5 & 78.0 & 54.0 & 66.0 & 73.0 & 72.4 & 72.8 \\
    CoF-sft Random Noise & 55.0 & 55.0 & 55.0 & 65.0 & 55.0 & 60.0 & 65.2 & 72.4 & 68.1 \\
    CoF-sft Ground Truth & 77.0 & 54.0 & 65.5 & 85.0 & 57.0 & 71.0 & 87.0 & 78.9 & 83.8 \\
    \bottomrule
    \end{tabular}
    \caption{Maximum visual tokens limited at 1,024.}
    \label{tab:my_awesome_results_1024}
    \end{subtable}

    \caption{Evaluation results with different cropped region settings.}
    \label{tab:my_awesome_results}

\end{table*}

To quantitatively evaluate the efficiency of existing models in utilizing identified cropped regions, we conducted experiments on three challenging high-resolution visual understanding datasets: HR-Bench-8k, HR-Bench-4k and $V^*$. For testing efficiency, we only retain one option order from four official option orders (marked as ``Lite") when testing on HR-Bench 8K and 4K. Following the inference process of previous work~\cite{zheng2025deepeyes,lai2025mini}, the input consists of a complete image and a corresponding question. The large language model can autonomously determine whether to invoke image cropping tools to assist in the answer. We limit the maximum number of visual tokens (obtained from MLLM’s vision encoder) for the complete input image at 16,384. This setting allows the full image to be processed at a high resolution. The cropped region are extracted from this original high-resolution image.  Building upon this, we further analyzed the model’s attention to cropped regions using the following two controlled settings: \textbf{1) Ground Truth Cropped Region}: Replacing the bounding boxes predicted by MLLM with ground truth bounding boxes creates a ``perfect" high-resolution cropped region. \textbf{2) Random Noise Cropped Region}: Replacing the high-resolution cropped region with random noise containing no useful information.

In both replacements, if the model truly pays attention to the provided cropped region, its performance should significantly improve when using the ground truth; conversely, its performance should drastically decrease when the cropped region is replaced with random noise.

As shown in Table \ref{tab:my_awesome_results_16384}, when we replace the MLLM's predicted bounding box with the ground truth or substitute the cropped region with random noise, the performance change is marginal (particularly for DeepEyes). This suggests that when the full input image retains rich information, the model tends to rely on the full image to answer questions, rather than the fine-grained details within the cropped region. We attribute this to the following: during training, the resolution of the full image and the source image for cropping are kept consistent. Consequently, the cropped region, compared to the full image, merely removes irrelevant context without providing any additional information. As a result, the MLLM fails to effectively learn to utilize the cropped region for answering questions during training, leading it to continue relying on the full image.

To further investigate whether introducing a resolution disparity at test time could encourage the model to focus on the cropped region, we downsample the full input image to a maximum of 1,024 tokens, while the cropped region are still extracted from the original, high-resolution image. This significant downsampling results in the loss of some fine details crucial for answering questions in these datasets. However, it preserves enough coarse information to allow the model to still locate the approximate region of interest.  Under this setting, we re-apply the two aforementioned replacement strategies. The results, presented in Table \ref{tab:my_awesome_results_1024}, show that although the performance change is more pronounced than in Table \ref{tab:my_awesome_results_16384}, however it still remains insignificant. This indicates that even when the information in the full image is insufficient, the model still fails to effectively attend to the cropped region.

We argue that the reason for this phenomenon is intuitive. In previous methods, the original image fed to model and the image used to obtain the cropped region shared the same resolution during training. Consequently, the introduction of the cropped region provided no additional information. As a result, the model failed to effectively learn how to leverage the cropped region to aid in reasoning.

Inspired by the above analysis, we introduce an informa-
tion gap mechanism during the first stage of training to explicitly encourage model to better focus on and utilize the cropped regions. Building upon this foundational ability, our second training stage is dedicated to enhancing model’s cropping precision to achieve further performance gains.

\section{Method}
\label{sec:method}

\begin{figure*}[t]
\centering
\includegraphics[width=\linewidth]{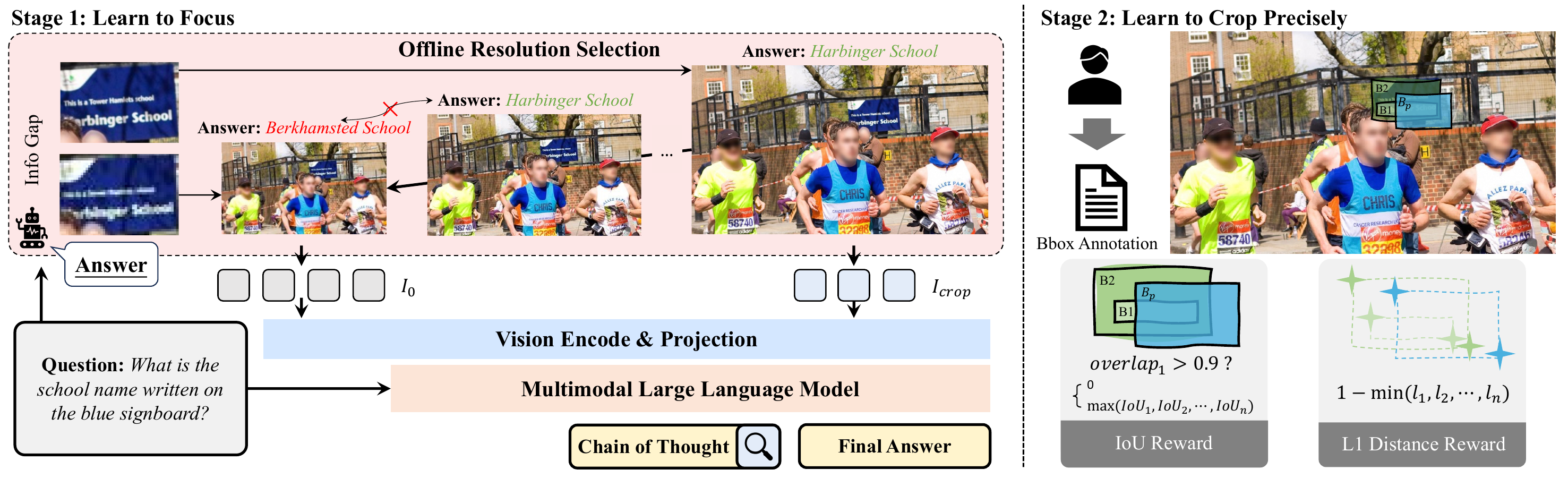} 
\caption{Framework of the proposed two-stage training method.} 
\label{frakework}
\end{figure*}

Consistent with prior work, we expect the model can predict the coordinates of a critical region, which is then cropped and leveraged to answer the given question. The action for the step $t$ can be fromulated as:
\begin{equation}
a_t \sim \pi_{\theta}(a \mid I_0, q, [r_1, I_1^{crop}], \dots, [r_m, I_m^{crop}]),
\end{equation}
where $I_0$ is the full input image, $r_i$ is the text response and $I_i^{crop}$ is the cropped image.
Our model is trained via a two-stage reinforcement learning process. In the first stage, we enhance the model's attention to the cropped region by introducing an information gap mechanism. In the second stage, we improve the model's cropping precision by annotating a small number of bounding boxes and incorporating a grounding reward.

\subsection{Learn to Focus}
The primary objective of our first training stage is to cultivate the model's ability to focus its attention on a specific cropped region for detailed analysis. We hypothesize that the information contained within both the complete input image, which we denote as $I_0$, and the region, $I_i^{crop}$, is crucial for this learning process. The complete image provides essential context, while the cropped region contains the fine-grained details necessary to answer the question.

To effectively train the model's focus, the cropped region must retain maximum detail, ensuring that the correct answer can be derived from it, thereby highlighting its importance. However, a critical challenge arises from the information balance in the complete image. If the complete image is excessively rich in detail (i.e., high-resolution), the model may develop a ``shortcut" learning behavior. It tends to directly extract the answer from the comprehensive global view rather than the crop (which is an issue that DeepEyes suffers from, as revealed in our  preliminary analysis). Conversely, if the complete image contains insufficient information, the model cannot identify the region relevant to the question, which renders the problem unsolvable. 

To strike this delicate balance, we introduce an Offline Resolution Selection process to prepare the optimal complete image for training. In this procedure, we start with an original high-resolution image from the training set. We then iteratively downsample it, creating a sequence of progressively lower-resolution versions. At each step, we use the to be trained MLLM to answer the question based on the original image and the current downsampled version. The process terminates as soon as the answer derived from the downsampled image diverges from the answer obtained from the original image. The image from the resoluton version that causes this divergence is selected as the optimal complete image for our training stage. This ensures it is just informative enough to provide context but not so detailed as to allow the model to bypass the high-resolution crop. Through this selection strategy, we formulate the training data for this stage. Each data instance is a tuple $(I_0, I_{ori}, q, answer)$, where $I_0$ is the selected optimal-resolution image, $I_{ori}$ is the original high-resolution image used for extracting the detailed cropped region $I_i^{crop}$, $q$ is the question, and $answer$ is the ground-truth response. 

In this stage, we formulates the total reward $r$ for a output trajectory as a sum of three distinct components:
\begin{equation}
r = r_{acc} + r_{format} + \mathbb{I}_{r_{acc}>0} \cdot r_{tool}.
\end{equation}
The accuracy reward is determined by whether the model response matches the ground-truth response, whereas the formatting reward penalizes outputs with deficient structures. Furthermore, a tool usage bonus is granted exclusively when the model not only provides a correct solution but also leverages at least one external perception tool during its process. We then compute the advantages $A$ by
normalizing the rewards and update the policy by adopting the GRPO \cite{shao2024deepseekmath, guo2025deepseek} optimization objective over mini-batches. Formally, the optimization objective is given by:
%\begin{equation}
    \begin{align}
    \mathcal{J} &= \mathbb{E}_{q \sim \mathcal{D}, \{o_i\}_{i=1}^N \sim \pi_{\theta'}(\cdot|q)} \Biggl[ \frac{1}{N} \sum_{i=1}N \biggl( \min \Bigl( \frac{\pi_{\theta}(o_i|q)}{\pi_{\theta'}(o_i|q)} A_i, \nonumber \\
    & \qquad \text{clip} \left( \frac{\pi_{\theta}(o_i|q)}{\pi_{\theta'}(o_i|q)}, 1 - \epsilon, 1 + \epsilon \right) A_i \Bigr) \biggr) \Biggr],
    \end{align}
%\end{equation}
where $A_i = \frac{r_i - \text{mean}(\{r_1, r_2, \dots, r_N\})}{\text{std}(\{r_1, r_2, \dots, r_N\})}.$

\subsection{Learn to Crop Precisely}
While the initial training stage enables the model to locate question-relevant regions, we observe a significant challenge stemming from the inherent information discrepancy in our approach. Specifically, the Multi-modal Large Language Model (MLLM) receives a down-sampled, low-resolution version of the full image, whereas the cropped patch retains high-resolution details. This asymmetry incentivizes the model, guided by the accuracy reward ($r_{acc}$), to progressively enlarge the predicted bounding box ($B_p$). A larger crop naturally contains more detailed information, making it easier for the MLLM to answer the question correctly and maximize its reward.

However, this behavior leads to two primary issues. First, the resulting oversized crops often include substantial redundant information, imposing unnecessary computational overhead on the MLLM and reducing overall efficiency. Second, this extraneous content can act as noise, potentially distracting the MLLM and causing interference during the reasoning process. To mitigate this, we introduce a second training stage focused on refining the agent's cropping policy with direct grounding supervision.

\noindent \textbf{Hierarchical Bbox Annotation.} We introduce an additional reward signal based on a small set of human-annotated bounding boxes. Inspired by the observation that visual regions of interest often possess a compositional and hierarchical structure \cite{kirillov2023segment, ravi2024sam, zhang2023personalize}, we annotate a series of nested ground-truth (GT) bboxes for each question-image pair, denoted as $B_1,B_2,…,B_n$. As illustrated in Figure 3 (right), these boxes range from the minimal necessary region required to answer the question ($B_1$, e.g., the text ``Harbinger School") to a larger, more contextually relevant area ($B_2$, e.g., the entire sign). This hierarchical annotation provides a more flexible and robust target for our model.

\noindent \textbf{Grounding Reward Formulation.} To guide the agent towards generating a precise yet sufficient crop, we design a composite grounding reward $r_{geo}$, which consists of an IoU reward and an $L_{1}$ distance reward.
The IoU Reward ($r_{IoU}$) encourages the predicted box $B_p$ to have a high overlap with one of the GT boxes. However, naively maximizing the Intersection over Union (IoU) could encourage the model to predict tighter bounding boxes, which may not completely cover the smallest GT box (e.g., $B_1$), thus causing the omission of crucial details needed to answer the question. To prevent this, we introduce a conditional reward. We first calculate the coverage ratio of $B_1$ by $B_p$, which we term $overlap$. The IoU reward is only granted if this ratio exceeds a threshold (set to 0.9 in our experiments). The reward is formulated as:
\begin {equation}
r_{IoU} = 
\begin{cases} 
\max_{i \in \{1,..,n\}} \text{IoU}_{i} & \text{if } overlap > \tau \\
0 & \text{otherwise}
\end{cases}
\end {equation}
where $IoU_{i} = \frac{\text{Area}(B_p \cap B_1)}{\text{Area}(B_p \cup B_1)}$
and $overlap = \frac{\text{Area}(B_p \cap B_1)}{\text{Area}(B_1)}.$
The condition of $r_{IoU}$ can lead to sparse rewards, especially early in training when the agent struggles to satisfy the overlap condition. To provide a denser and more consistent training signal, we incorporate an $L_{1}$ Distance Reward ($r_{l1}$). This reward measures the normalized $L_{1}$ distance between the corners of the predicted box and the closest GT box, ensuring the agent receives a corrective gradient even when the IoU reward is zero. The $L_{1}$ reward is defined as:
\begin{equation}
r_{l1} = 1 - \min_{i \in \{1,..,n\}} d_{L_{1}}(B_p, B_i)
\end{equation}
Finally, the grounding reward is the combination of the IoU reward and the $L_{1}$ reward:
\begin{equation}
r_{geo} = \omega*r_{IoU} + (1 - \omega)*r_{l1}.
\end{equation}

\section{Experiment}
\label{sec:exp}

\subsection{Setups}

\begin{figure}[t]

\centering
\includegraphics[width=0.8\linewidth]{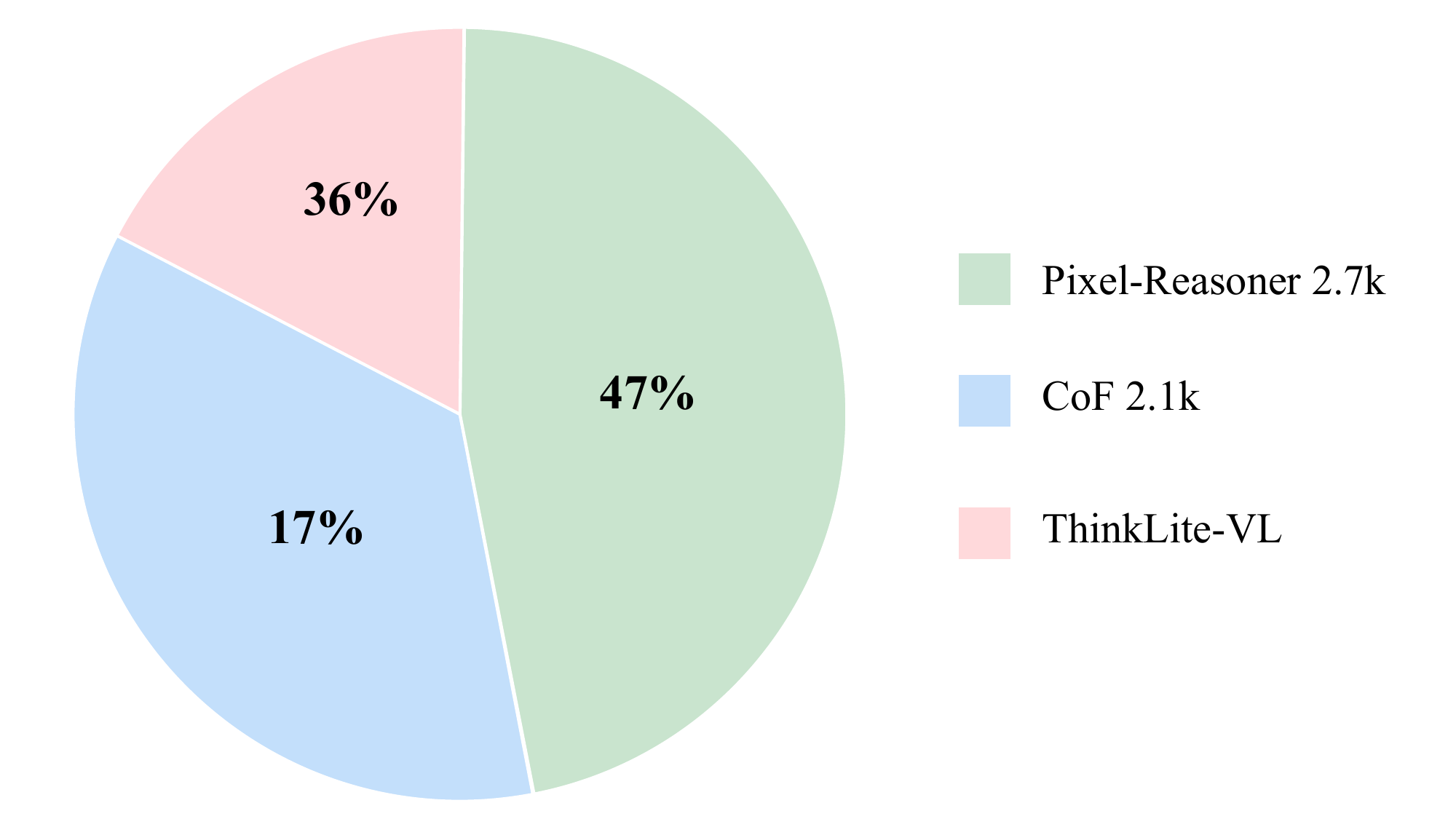} 
\caption{Training data distribution in Stage-I}
\label{fig:data_dis}
\end{figure}

\textbf{Training Details}. We train Qwen2.5-VL-7B-Instruct on 8 A100 GPUs for 80 steps with GRPO. Each step contains 256 samples and 16 rollouts per sample. The maximum response length is set to 2048. The learning rate is $1 \times 10^{-6},$ and neither KL regularization nor entropy is applied.

\noindent\textbf{Training Datasets.} In the first stage, we address the potential information loss by down-sampling high-resolution images. To this end, we train the model's visual perception capabilities using high-resolution VQA data from Pixelreasoner and CoF, and enhance its analytical abilities with a small set of samples from ThinkLite-VL. The distribution of the datasets is detailed in Figure ~\ref{fig:data_dis}. Notably, while Pixelreasoner and CoF also provide reasoning trajectories generated by MLLMs for supervised fine-tuning, our training methodology does not utilize them. We only require the images and their corresponding question-answer (QA) pairs.
We observed that after the first stage of training, the model could already crop regions that generally covered the question-relevant areas. To refine this capability and further improve the cropping precision, we curated a dataset for the second stage by sampling 256 instances from VisualProbe datasets from Mini-o3 \cite{lai2025mini} and manually annotating them with precise bounding boxes (BBoxes).

\begin{table*}[t] % [H] 强制表格出现在此处; 您也可以用 [htbp] 让 LaTeX 自动选择位置
    \centering % 将整个表格（包括子表）居中

    \begin{subtable}{0.9\textwidth}
    \centering
    \small % 使用稍小的字体以确保表格能优雅地放入页面
    \setlength{\tabcolsep}{4.5pt} % 微调列之间的水平间距，可根据需要调整
    \label{tab:main_results}
    \begin{tabular}{@{}ccc|ccc|ccc|ccc@{}}
    \toprule
    \multirow{2}{*}{\textbf{Model}} & \multirow{2}{*}{\textbf{Size}} & \multirow{2}{*}{\textbf{Trajectory Free}} & \multicolumn{3}{c}{HR-Bench 8K} & \multicolumn{3}{c}{HR-Bench 4K} & \multicolumn{3}{c}{$V^{*}$} \\
    \cmidrule(lr){4-6} \cmidrule(lr){7-9} \cmidrule(lr){10-12}
    & & & \textbf{FSP} & \textbf{FCP} & \textbf{Overall} & \textbf{FSP} & \textbf{FCP} & \textbf{Overall} & \textbf{Attr} & \textbf{Spatial} & \textbf{Overall} \\
    \midrule
    CoF-sft \cite{zhang2025chain} & 7B & $\times$   & 66.0 & 51.7 & 58.9 & 75.5 & 56.5 & 66.0 & 73.0 & 72.4 & 72.8 \\
    Pixel Reasoner \cite{wang2025pixel} & 7B  & $\times$  & 62.8 & 57.0 & 59.9 & 69.8 & 61.8 & 65.8 & 73.1 & 75.0 & 73.9 \\
    Mini-o3 \cite{lai2025mini}  & 7B & $\times$   & 77.5 & 54.5 & 66.0 & 82.0 & 59.3 & 70.7 & 78.3 & \textbf{82.9} & 80.1 \\
    \midrule
    DeepEyes \cite{zheng2025deepeyes} & 7B & \checkmark & 68.8 & 55.2 & 62.0 & 73.0 & 54.7 & 63.9 & 75.7 & 76.3 & 75.9 \\
    \midrule
    \textbf{Ours}     & \textbf{7B} & \textbf{\checkmark} & \textbf{83.3} & \textbf{60.8} & \textbf{72.1} & \textbf{86.8} & \textbf{63.7} & \textbf{75.3} & \textbf{82.6} & 77.6 & \textbf{80.6} \\
    \bottomrule
    \end{tabular}
      \caption{Maximum visual tokens limited at 1,024.} % 表格标题
      \label{tab:compare_1024} % 用于在文中引用表格，例如 \ref{tab:main_comparison}
    \end{subtable}

    \begin{subtable}{0.9\textwidth}
    \centering
    \begin{threeparttable}
    \small % 使用稍小的字体以确保表格能优雅地放入页面
    \setlength{\tabcolsep}{4.5pt} % 微调列之间的水平间距，可根据需要调整
    \label{tab:main_results}
    \begin{tabular}{@{}ccc|ccc|ccc|ccc@{}}
    \toprule
    \multirow{2}{*}{\textbf{Model}} & \multirow{2}{*}{\textbf{Size}} & \multirow{2}{*}{\textbf{Trajectory Free}} & \multicolumn{3}{c}{HR-Bench 8K} & \multicolumn{3}{c}{HR-Bench 4K} & \multicolumn{3}{c}{$V^{*}$} \\
    \cmidrule(lr){4-6} \cmidrule(lr){7-9} \cmidrule(lr){10-12}
    & &                                                                             & \textbf{FSP}   & \textbf{FCP}   & \textbf{Overall} & \textbf{FSP}   & \textbf{FCP}   & \textbf{Overall} & \textbf{Attr}  & \textbf{Spatial} & \textbf{Overall} \\
    \midrule
    CoF-sft \cite{zhang2025chain}                 & 7B & $\times$                     & 87.7  & 51.0  & 69.4    & 92.2  & 53.7  & 73.0    & 90.4  & 85.5    & 88.5 \\
    Pixel Reasoner \cite{wang2025pixel}           & 7B & $\times$                     & 78.3  & 55.5  & 66.9    & 85.8  & 61.5  & 73.6    & 82.6  & 86.8    & 84.3 \\
    Mini-o3 \cite{lai2025mini}                  & 7B & $\times$                     & 82.0  & 49.2  & 65.6    & 84.0  & 54.0  & 69.0    & 87.8  & \textbf{88.2}    & 88.0 \\
    Mini-o3$^\dagger$ \cite{lai2025mini}        & 7B & $\times$                     & \textcolor{lightgray}{-}     &\textcolor{lightgray} {-}     & \textcolor{lightgray} {73.3}    & \textcolor{lightgray}{-}     & \textcolor{lightgray}{-}     & \textcolor{lightgray}{77.5}    & \textcolor{lightgray}{-}     & \textcolor{lightgray}{-}       & \textcolor{lightgray}{88.2} \\
    \midrule
    DeepEyes \cite{zheng2025deepeyes}           & 7B & \checkmark                   & 84.3  & 54.7  & 69.5    & 90.5  & 55.2  & 72.9    & 84.3  & 88.2    & 85.9 \\
    DeepEyes$^\ddagger$\cite{zheng2025deepeyes} & 7B & \checkmark                   & \textcolor{lightgray}{86.8}  & \textcolor{lightgray}{58.5}  & \textcolor{lightgray}{72.6}    & \textcolor{lightgray}{91.3}  & \textcolor{lightgray}{59.0}  & \textcolor{lightgray}{75.1}    & \textcolor{lightgray}{91.3}  & \textcolor{lightgray}{88.2}    & \textcolor{lightgray}{90.1} \\
    \midrule
    \textbf{Ours}     & 7B & \textbf{\checkmark}                                    & \textbf{87.7}  & \textbf{63.0}  & \textbf{75.4}    & \textbf{92.7}  & \textbf{60.0}  & \textbf{76.4}    & \textbf{91.3}  & 86.8    & \textbf{89.5} \\
    \bottomrule
    \end{tabular}
    \begin{tablenotes}
      \footnotesize % 可以设置脚注字体大小
      \item[$\dagger$] Result from Mini-o3 paper, where it was obtained by setting temperature to 1.0 and averaging results over 32 runs for each question.
      \item[$\ddagger$] Result from DeepEyes paper, where they did not shuffle the order of options on the $V^{*}$ dataset, unlike our evaluation.
    \end{tablenotes}
       
      \caption{Maximum visual tokens limited at 16,384.} % 表格标题
      \label{tab:compare_16384} % 用于在文中引用表格，例如 \ref{tab:main_comparison}
    \end{threeparttable}
    \end{subtable}

    \caption{Comparison with other models. Our method achieves SOTA performance on HR-Bench 8K, HR-Bench 4K and $V^{*}$ benchmarks.}
    \label{tab:SOTA_results}

\end{table*}

\noindent\textbf{Benchmarks and Compared Methods.}
Our test benchmark is built upon HR-Bench 8K, HR-Bench 4K and $V^{*}$. To evaluate a model's ability to answer questions based on cropped regions, we limit the maximum number of visual tokens to 1,024 and 16,384 respectively. However, the cropped regions themselves are extracted from the original, uncompressed images. The models we compare against include CoF (only the SFT-stage model is publicly available), Pixel Reasoner and Mini-o3, which are trained via a two-stage process of SFT+RL, and DeepEyes, which is trained purely with RL.
For the evaluation on Mini-o3, we disabled sampling and employed a greedy decoding strategy to maintain consistency with the testing setups of other models. Furthermore, we find that Mini-o3 exhibits a propensity for a high number of calls to the cropping tool. To address this, we imposed an upper limit of 12 dialogue turns. If the model reached this maximum turn count without producing a final answer, we classified the attempt as incorrect.

\subsection{Results}

\begin{table*}[t]
  \centering

  % 减小列间距以适应页面宽度，可根据你的文档边距微调
  \setlength{\tabcolsep}{4pt} 
  \begin{tabular}{l cc cc c cc cc c cc cc c}
    \toprule
    % \multirow{3}{*}{Model} 创建了一个跨3行的单元格来垂直居中“Model”
    \multirow{3}{*}{\textbf{Model}} & \multicolumn{5}{c}{HR-Bench 8k} & \multicolumn{5}{c}{HR-Bench 4k} & \multicolumn{5}{c}{$V^{*}$} \\
    % \cmidrule 用于创建跨越部分列的水平线
    \cmidrule(lr){2-6} \cmidrule(lr){7-11} \cmidrule(lr){12-16}
    & \multicolumn{2}{c}{\textbf{FSP}} & \multicolumn{2}{c}{\textbf{FCP}} & \multirow{2}{*}{\textbf{Acc}} & \multicolumn{2}{c}{\textbf{FSP}} & \multicolumn{2}{c}{\textbf{FCP}} & \multirow{2}{*}{\textbf{Acc}} & \multicolumn{2}{c}{\textbf{Attr}} & \multicolumn{2}{c}{\textbf{Spatial}} & \multirow{2}{*}{\textbf{Acc}} \\
    \cmidrule(lr){2-3} \cmidrule(lr){4-5} \cmidrule(lr){7-8} \cmidrule(lr){9-10} \cmidrule(lr){12-13} \cmidrule(lr){14-15}
    & \textbf{Acc} & \textbf{IoU} & \textbf{Acc} & \textbf{IoU} & & \textbf{Acc} & \textbf{IoU} & \textbf{Acc} & \textbf{IoU} & & \textbf{Acc} & \textbf{IoU} & \textbf{Acc} & \textbf{IoU} & \\
    \midrule
    Baseline     & 65.2          & 28.5 & 56.8           & 40.1 & 61.0 & 75.0 & 30.3 & 60.8 & 45.6 & 67.9 & 72.2 & \textbf{39.1} & 72.4 & 30.3 & 72.3 \\
    Stage-I   & 77.7 & 29.8          & \textbf{62.0} & 39.3           & 69.9 & 85.3 & 32.3 & 63.5 & 46.5 & 74.4 & 80.0 & 25.7 & 75.0 & 21.0 & 78.0 \\
    %+ Stage 2    & 76.7 & 47.5          & 59.5 & 56.2           & 68.1 & 82.5 & 45.8 & 62.0 & 57.0 & 72.3 & 79.1 & 43.6 & 75.0 & 43.9 & 77.5 \\
    %+ VP Data    & 81.8          & 39.3 & \textbf{63.5}           & 47.4 & \textbf{72.7} & 83.8 & 39.2 & 62.0 & 53.0 & 72.9 & 77.4 & 37.5 & 75.0 & 29.2 & 76.4 \\
    Stage-II &\textbf{83.3} & \textbf{34.5} & 60.8  & \textbf{44.4} & \textbf{72.1} &\textbf{86.8} & 37.8 &\textbf{63.7} & \textbf{51.0} &\textbf{75.3} &\textbf{82.6} & 37.4 &\textbf{77.6} & \textbf{30.8} & \textbf{80.6} \\
    \bottomrule
  \end{tabular}
  \caption{Effectiveness of two-stage training strategy. The best result is indicated in \textbf{bold}.}
  \label{tab:ablation_1}
\end{table*}

\begin{table*}[htbp]
  \centering

  % 减小列间距以适应页面宽度，可根据你的文档边距微调
  \setlength{\tabcolsep}{4pt} 
  \begin{tabular}{l c c c c c c c c c}
    \toprule
    % \multirow{3}{*}{Model} 创建了一个跨3行的单元格来垂直居中“Model”
    \multirow{2}{*}{\textbf{Model}} & \multicolumn{3}{c}{HR-Bench 8k} & \multicolumn{3}{c}{HR-Bench 4k} & \multicolumn{3}{c}{$V^{*}$} \\
    % \cmidrule 用于创建跨越部分列的水平线
    \cmidrule(lr){2-4} \cmidrule(lr){5-7} \cmidrule(lr){8-10}
    & \textbf{FSP} & \textbf{FCP} & \textbf{Overall} & \textbf{FSP} & \textbf{FCP} & \textbf{Overall} & \textbf{Attr} & \textbf{Spatial} & \textbf{Overall} \\
    \midrule
    Stage-I   & 77.7 & 62.0 & 69.9 & 85.3 & 63.5 & 74.4 & 80.0 & 75.0 & 78.0 \\
    \midrule
    + Original Data  & 76.7 & 59.5 & 68.1 & 82.5 & 62.0 & 72.3 & 79.1 & 75.0 & 77.5 \\
    + VP Data & 81.8 & \textbf{63.5} & \textbf{72.7} & 83.8 & 62.0 & 72.9 & 77.4 & 75.0 &  76.4 \\
    + VP Data \& L1 Reward  &\textbf{83.3} & 60.8  & 72.1 &\textbf{86.8} &\textbf{63.7} &\textbf{75.3} &\textbf{82.6} &\textbf{77.6}& \textbf{80.6} \\
    \bottomrule
  \end{tabular}
  \caption{Effectiveness of different designs on the second training stage. Original Data means training the model with the data from the same source used in the first stage. VP Data denotes the VisualProb datasets. The setting of VP Data \& L1 Reward is denoted as Stage-II.}
  \label{tab:ablation_3}
\end{table*}

\begin{table}[t]
    \centering
    \begin{tabular}{cccc}
    \toprule
     \textbf{Model}    & HR-Bench 8k & HR-Bench 4k & $V^{*}$ \\
    \midrule
      Baseline   & 4.5 & 0.5 & 3.2 \\
    \midrule
    Stage-I & 18.5 & 10.5 & 11.0 \\
    \bottomrule
    \end{tabular}
    \caption{Acc delta of replacing the cropped images with ground truth or random noise on the three benchmarks.}
    \label{tab:abltion_2}
\end{table}

As shown in Table \ref{tab:SOTA_results}, our method achieves state-of-the-art results on the evaluated benchmarks when the number of visual tokens is limited to both 1,024 and 16,384. Notably, under the 1024-token constraint, our method significantly outperforms other approaches, and this performance advantage becomes more pronounced as the input image resolution increases. This phenomenon indicates that our method can effectively utilize information from the cropped regions to aid inference and demonstrates a significant performance advantage in high-resolution scenarios.

\subsection{Ablation Study}

We conduct a comprehensive ablation study to validate the effectiveness of each component. All ablation studies in this section are conducted with the maximum number of visual tokens set to 1,024 if not specified.

\noindent\textbf{Two-Stage Training.} We first validate the effectiveness of the two-stage training strategy. As summarized in Table~\ref{tab:ablation_1}, compared to training by following earlier work~\cite{zheng2025deepeyes,lai2025mini} (denoted as the baseline), the information gap we introduce in the first stage of training effectively helps the model quickly learn to utilize detailed information in the cropped image, thereby improving the accuracy of the responses. On the HR-Bench 8k, HR-Bench 4k, and $V^{*}$ test sets, the average accuracy improved by 8.9\%, 6.5\%, and 5.7\%, respectively. 
To further demonstrate the model's attention to the cropped image, we summarize the changes in model response accuracy after replacing the cropped image in Table~\ref{tab:abltion_2}, referring to the testing method in Section~\ref{sec:formatting}. Clearly, the model trained in the first stage pays more attention to information from the cropped image. In the second stage, we focus on training the model's ability to accurately locate and crop local details. Table~\ref{tab:ablation_1} also shows that this stage of training can further improve the accuracy of the model's responses and achieve a significant improvement in the IoU metric for cropped images (around 5\%-9\%).

\noindent\textbf{Effect of Data Selecting.} As shown in Table~\ref{tab:ablation_3}, an intuitive approach in the second stage of training is to select a portion of the training data from the first stage, label it, and then train and optimize the model. However, we find that the model trained in the first stage can correctly answer almost all the questions on the training data, and direct training does not bring any additional gains; in fact, it even slightly decreased in the three benchmark tests. Therefore, we chose the more challenging VisualProb datasets \cite{lai2025mini} for the second stage of training. With the help of the IoU reward, the model improves its accuracy by an average of 4.6\% in the most challenging benchmark test, HR-Bench 8k.

\noindent\textbf{Effect of L1 Reward.} While using VisualProb data can help the model improve response accuracy under high-resolution conditions, more difficult learning samples increase the sparsity of IoU rewards, reducing the model's learning efficiency. Therefore, we additionally add an $L_{1}$ reward for the location center distance to help the model learn how to locate more accurately in difficult samples. The results in Table~\ref{tab:ablation_3} show that the introduction of this reward mechanism can effectively improve the response accuracy on all three test benchmarks, significantly exceeding the performance of the Stage-I model. 

\section{Efficiency Analysis}

We compare the accuracy (Acc) and inference time per question (Time) of our method against others. For this comparison, our method processes full images with a maximum of 1,024 visual tokens, whereas others utilize up to 16,384 tokens. The results, as shown in Table \ref{tab:efficiency}, indicate that our method surpasses existing approaches even when using lower-resolution complete images (implied by the fewer tokens). Due to the significantly smaller number of visual tokens, our method also demonstrates a substantial advantage in efficiency, with a much lower time cost per question. This indicates that our approach enables the model to utilize visual tokens more efficiently, which we attribute to its ability to focus on accurately cropped regions of interest.

\begin{table}[t]
    \centering
  \begin{tabular}{l c c c c}
    \toprule
    % \multirow{3}{*}{Model} 创建了一个跨3行的单元格来垂直居中“Model”
    \multirow{2}{*}{\textbf{Model}} & \multicolumn{2}{c}{HR-Bench 8k} & \multicolumn{2}{c}{HR-Bench 4k}  \\
    % \cmidrule 用于创建跨越部分列的水平线
    \cmidrule(lr){2-3} \cmidrule(lr){4-5} 
    & \textbf{Acc}  & \textbf{Time (s) $\downarrow$}  & \textbf{Acc}  & \textbf{Time (s) $\downarrow$}  \\
    \midrule
    CoF-sft                   & 69.4  & 6.0  & 73.0 & 5.5  \\
    Pixel Reasoner                & 66.9  & 5.0  & 73.6 & 8.4  \\
    Mini-o3                     & 65.6  & 27.8 & 69.0 & 21.4  \\
    DeepEyes                    & 69.5  & 12.4 & 72.9 & 9.3  \\
    \midrule
    \textbf{Ours$^{1,024}$}        & \textbf{72.1}  & \textbf{2.8}  & \textbf{75.3} & \textbf{2.6}  \\
    \bottomrule
  \end{tabular}
  \caption{Our method processes complete images with a maximum of 1,024 visual tokens, whereas the others utilize up to 16,384.}
  \label{tab:efficiency}
\end{table}

\section{Conclusion}
\label{sec:sum}

We analyze limitations of existing agent-based MLLMs, which overemphasize original input image while neglecting cropped region. To guide the model to utilize crop, we propose a two-stage RL training strategy. In first stage, we construct an information gap mechanism by compressing input image while maintaining the resolution of cropped region, forcing model to attend to details in the cropped image. In second stage, we select a smaller while challenging dataset and further optimize the model's ability to locate regions.

%\section{Acknowledgment}
%This work was supported by the Anhui Provincial %Science and Technology Major Project (No. %2023z020006), the Anhui Provincial Natural Science %Foundation (2508085QF212), and the advanced %computing resources provided by the Supercomputing %Center of the USTC.

{
    \small
    \bibliographystyle{ieeenat_fullname}
    \bibliography{main}
}
\clearpage
\setcounter{page}{1}
\maketitlesupplementary

\begin{figure*}[t]
    \centering % 居中整个 figure 环境
    
    \begin{subfigure}[b]{0.48\textwidth}
        \centering
        \includegraphics[height=5cm]{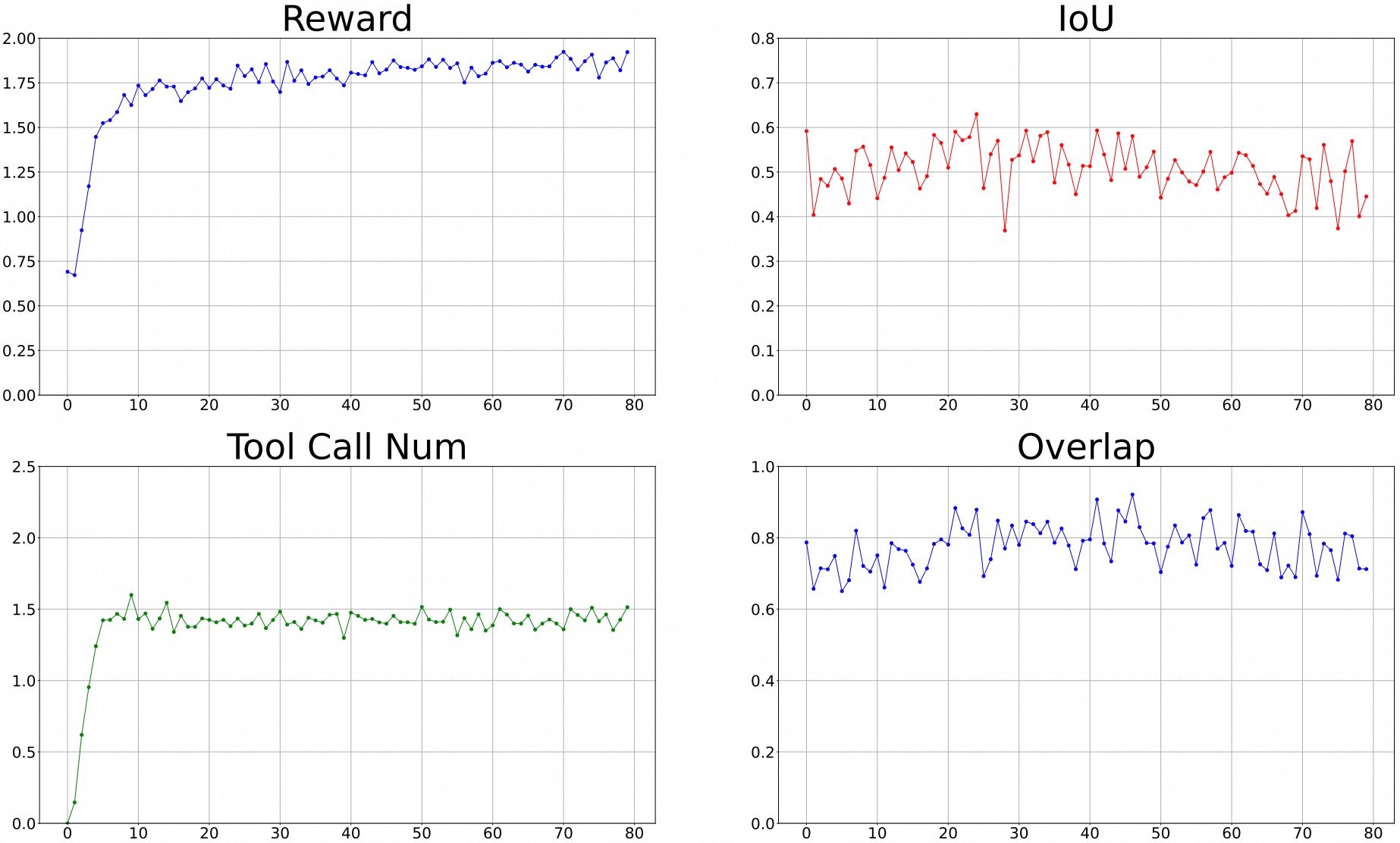}
        \caption{Traing progress of BaseLine}
        %\label{fig:progress_1}
    \end{subfigure}
    \hfill % 在两个子图之间添加弹性水平间距，使其自动分布在两端
    % 第二个子图
    \begin{subfigure}[b]{0.48\textwidth}
        \centering
        % 使用占位符图片 example-image-b，您需要替换成自己的图片路径
        % 关键点：设置与第一个子图相同的高度
        \includegraphics[height=5cm]{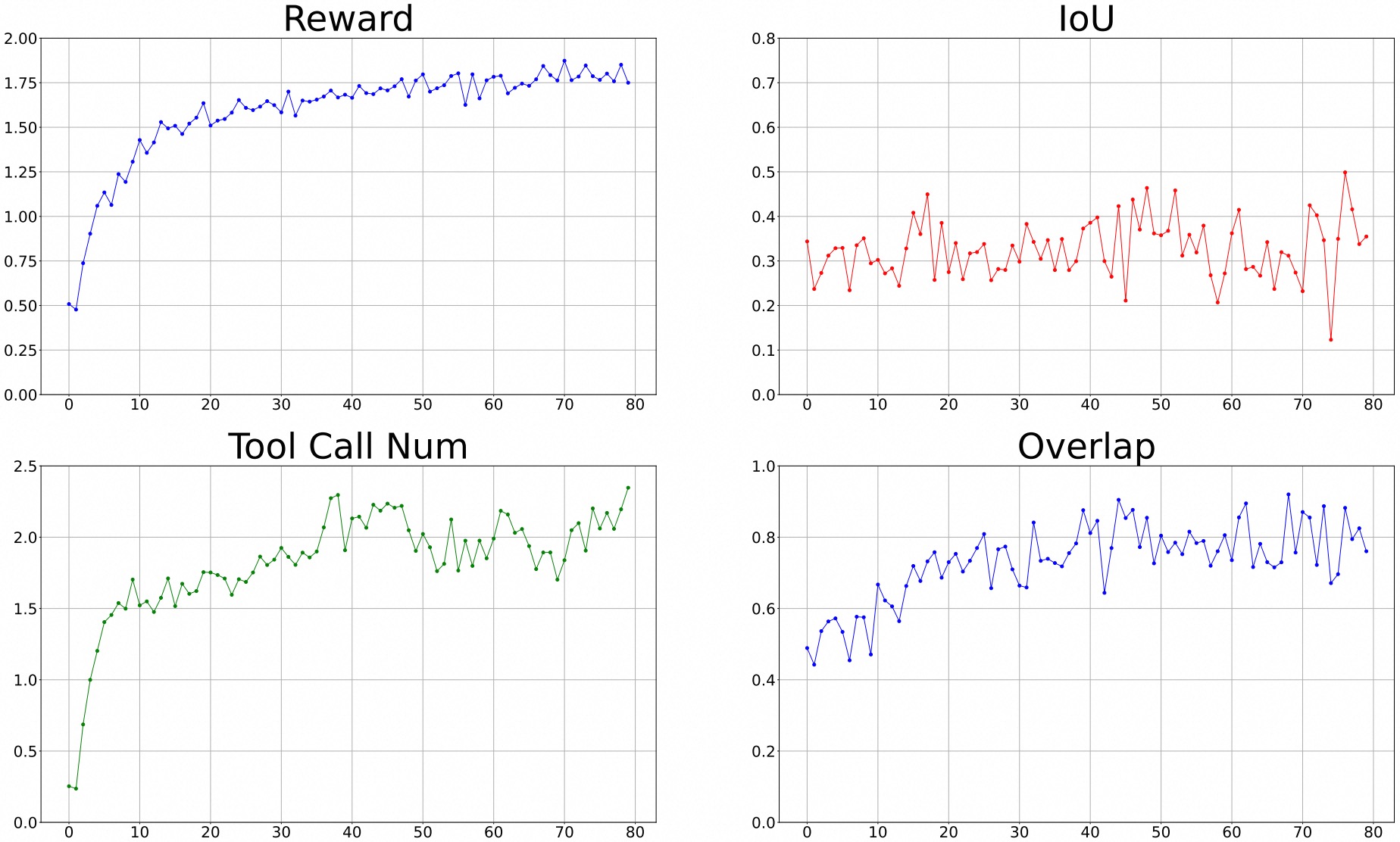}
        \caption{Traing progress with Info Gap}
        %\label{fig:progress_1}
    \end{subfigure}
    
    \caption{Training progress of BaseLine and Stage-I model.}
    \label{fig:progress_1}
\end{figure*}

\begin{figure*}[t]
    \centering % 居中整个 figure 环境
    
    \begin{subfigure}[b]{0.49\textwidth}
        \centering
        \includegraphics[height=7.5cm]{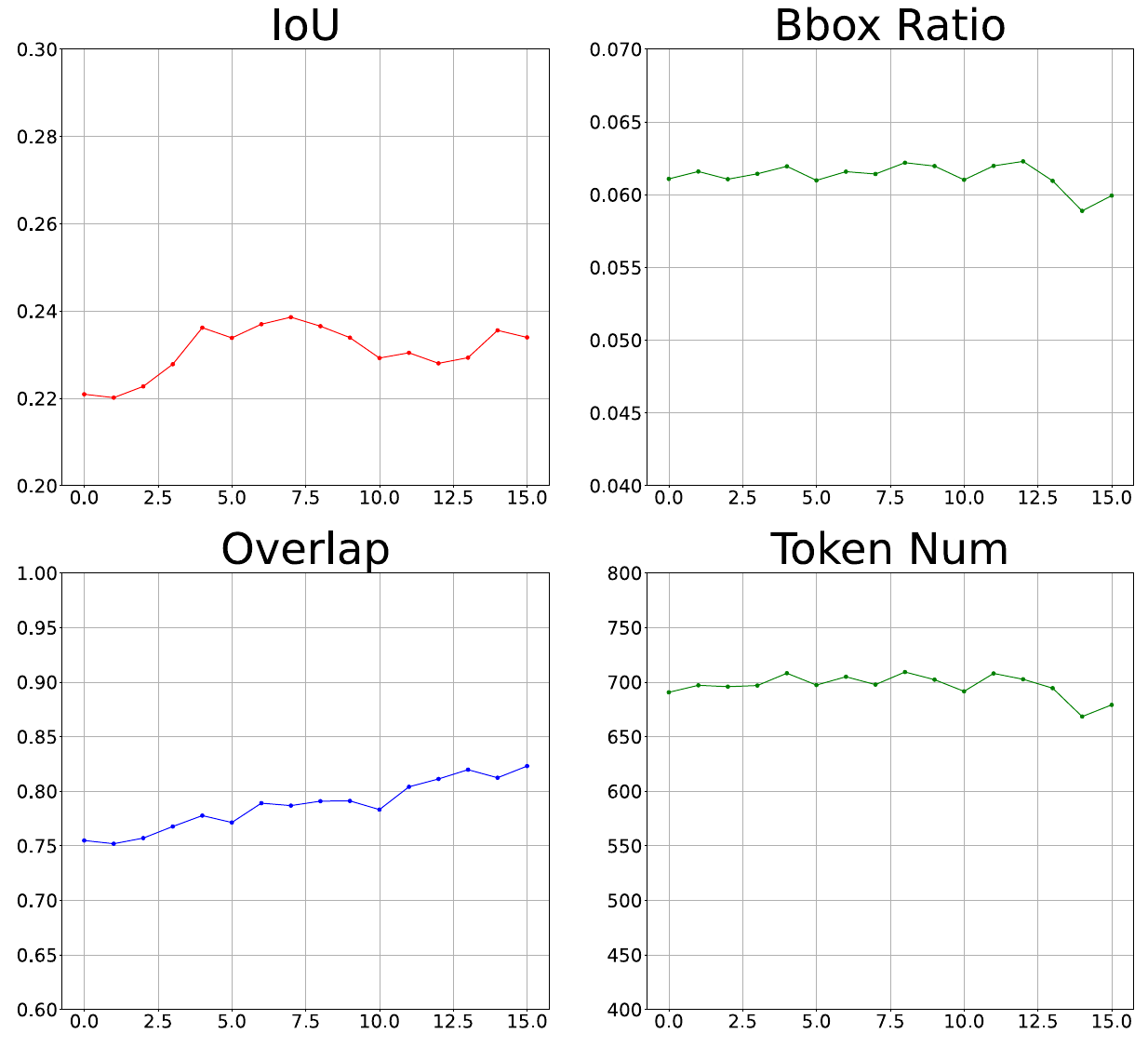}
        \caption{Training progress without grounding reward}
    
    \end{subfigure}
    \hfill % 在两个子图之间添加弹性水平间距，使其自动分布在两端
    % 第二个子图
    \begin{subfigure}[b]{0.49\textwidth}
        \centering
        % 使用占位符图片 example-image-b，您需要替换成自己的图片路径
        % 关键点：设置与第一个子图相同的高度
        \includegraphics[height=7.5cm]{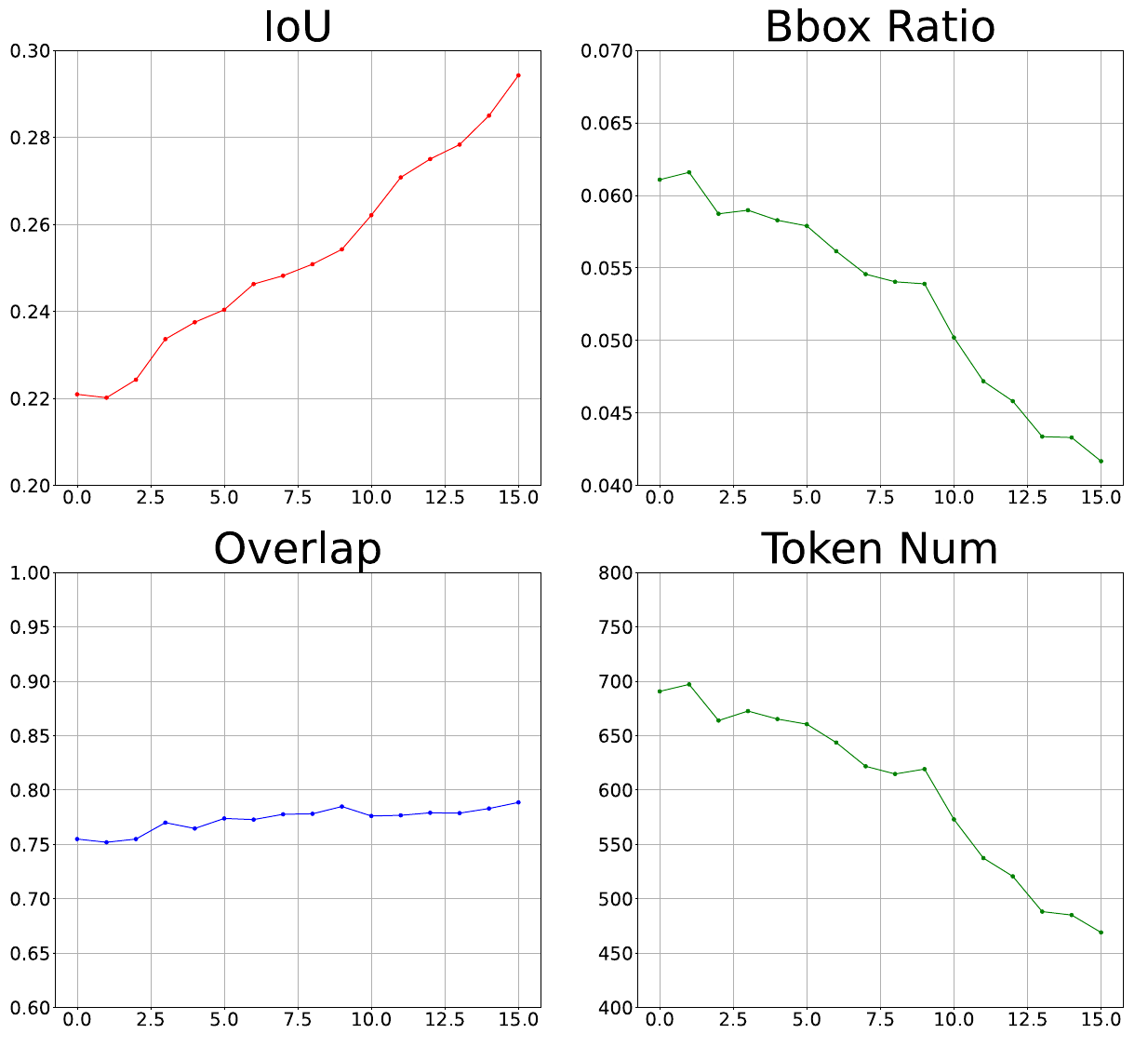}
        \caption{Traing progress with grounding reward}

    \end{subfigure}
    
    \caption{Training progress of Stage-II.}
    \label{fig:progress_2}
\end{figure*}

\section{Training Dynamics Analysis}
In this section, we visualize and analyze evolution of several key metrics during the first and second stages of training.
\subsection{Stage-I}
To analyze the model's behavior during the first stage of training, we present the evolution of four key metrics for both the BaseLine and our Stage-I Model with Info Gap in Figure~\ref{fig:progress_1}. These metrics are: the total reward, the number of tool calls (Tool Call Num), the Intersection over Union (IoU) of predicted crop regions with ground truth, and the overlap ratio (Overlap), which measures the proportion of ground-truth region covered by the model's predicted crops.

\noindent\textbf{Reward.} As shown in the `Reward' plots, the BaseLine model (Figure ~\ref{fig:progress_1} (a)) exhibits a faster initial increase in reward. We attribute this to uncompressed images in its training set, which present a simpler learning task. Critically, both models eventually converge to a similar reward level. This demonstrates that the introduction of the information gap compels Stage-I to achieve a comparable performance level on more challenging (compressed) data, demonstrating the model's effective utilization of the crop.

\noindent\textbf{Tool Calls.} The `Tool Call Num' for both models increases rapidly and then stabilizes. Notably, the Stage-I Model (Figure ~\ref{fig:progress_1} (b)) converges to a higher number of tool calls. We hypothesize this is because the compressed input image necessitates more exploratory crops to gather sufficient information to answer the question. This behavior confirms that the Stage-I Model learns to actively utilize the tool to attend to relevant image regions.

\noindent\textbf{IoU.} The `IoU' metric, which evaluates the precision of the cropped regions, shows no significant improvement for either model throughout the training process. This suggests that without an explicit grounding reward, the model is not incentivized to refine the precision of its crops. The consistently higher IoU of the BaseLine is likely due to the uncompressed images making box prediction an easier task.

\noindent\textbf{Overlap.} The `Overlap' reveals a distinct difference between the training progress of BaseLine and Stage-I. The BaseLine's overlap remains relatively stable without a clear upward trend. In contrast, the Stage-I's overlap shows a significant increase before plateauing. This indicates that while image compression initially makes it difficult for the model to identify question-relevant regions, the model learns a stronger region identification capability, driven by the accuracy reward, to correctly answer the question.

\subsection{Stage-II}
The Figure \ref{fig:progress_2} illustrates the evolution of four key metrics during the second training stage, with and without the use of the grounding reward. These metrics are: IoU, overlap, the ratio of the predicted crop box area to the image area (Bbox Ratio), and the visual token number of the predicted cropped region (Token Num).

\noindent\textbf{IoU.} 
Comparing the IoU evolution in Figures \ref{fig:progress_2} (a) and (b), we observe a distinct upward trend in (b) during Stage 2 training. This indicates that the introduction of the grounding reward enables the model to localize key regions with greater precision.

\noindent\textbf{Bbox Ratio and Token Num.} Comparison for the Bbox Ratio reveals that this metric remains largely constant in (a), whereas it decreases significantly in (b), leading to a subsequent reduction in the Token Num. This demonstrates that the grounding reward encourages smaller crop regions, thereby improving the model's inference efficiency.

\noindent\textbf{Overlap.} As shown in (b), the Overlap metric exhibits a moderate increase. This suggests that the model is not indiscriminately shrinking the crop area; rather, it is learning to prune redundant, non-critical parts of the region. Consequently, the resulting crop contains less distracting information, making it more conducive for the model to answer questions based on the visual content.

\begin{table}[h] % [H] 强制表格出现在此处; 您也可以用 [htbp] 让 LaTeX 自动选择位置
    \centering % 将整个表格（包括子表）居中

    \begin{subtable}{0.45\textwidth}
    \centering % 表格居中
    \begin{tabular}{lccc}
    \toprule
    \textbf{Setting} & HR-Bench 8K & HR-Bench 4K & $V^{*}$ \\
    \midrule
    prediction  & 41.7 & 47.0 & 55.6 \\
    GT          & 50.0 & 53.0 & 60.3 \\
    \bottomrule
    \end{tabular}
    \caption{GT Test.}
    \label{more_preliminary_gt} % 用于在正文中引用该表格，例如 
    \end{subtable}
    
    \begin{subtable}{0.45\textwidth}
    \centering % 表格居中
    \begin{tabular}{lccc}
    \toprule
    \textbf{Setting} & HR-Bench 8K & HR-Bench 4K & $V^{*}$ \\
    \midrule
    prediction     & 100.0 & 100.0 & 100.0 \\
    RandNoise   & 100.0 & 80.8 & 100.0 \\
    \bottomrule
    \end{tabular}
    \caption{Noise test.}
    \label{more_preliminary_noise} % 用于在正文中引用该表格，例如 
    \end{subtable}

    \caption{Results for more rigorous preliminary analysis.}
    \label{more_preliminary}

\end{table}

\begin{table*}[t]
  \centering

  % 减小列间距以适应页面宽度，可根据你的文档边距微调
  \setlength{\tabcolsep}{4pt} 
  \begin{tabular}{l c c c c c c c c c}
    \toprule
    % \multirow{3}{*}{Model} 创建了一个跨3行的单元格来垂直居中“Model”
    \multirow{2}{*}{\textbf{Model}} & \multicolumn{3}{c}{HR-Bench 8k} & \multicolumn{3}{c}{HR-Bench 4k} & \multicolumn{3}{c}{$V^{*}$} \\
    % \cmidrule 用于创建跨越部分列的水平线
    \cmidrule(lr){2-4} \cmidrule(lr){5-7} \cmidrule(lr){8-10}
    & \textbf{FSP} & \textbf{FCP} & \textbf{Overall} & \textbf{FSP} & \textbf{FCP} & \textbf{Overall} & \textbf{Attr} & \textbf{Spatial} & \textbf{Overall} \\
    \midrule
    Mix Data   & 77.5 & 59.0 & 68.3 & 85.0 & 61.0 & 73.0 & 82.6 & 73.7 & 79.1 \\
    Stage-II  & 83.3 & 60.8 & 72.1 & 86.8 & 63.7 & 75.3 & 82.6 & 77.6 & 80.6 \\
    \bottomrule
  \end{tabular}
  \caption{Comparison between different data utilization strategies. Mix Data: Using the mixture of our collected data and Visual Probe data to train the model in a single stage. Stage-II: First, train the model on our collected data, and then train it with a minimal amount of Visual Probe data with grounding reward.}
  \label{tab:ablation_data}
\end{table*}

\begin{table}[t]
    \centering
    \begin{tabular}{cccc}
    \toprule
     \textbf{Model}             & \textbf{HR-Bench 8k}    & \textbf{HR-Bench 4k}     & \textbf{V Star} \\
    \midrule
      Hard             & 60.5           & 63.2            & 71.1 \\
      Random           & 59.0           & 64.0            & 72.4 \\
    \midrule
      Answer     & 62.0           & 63.5            & 75.0 \\
    \bottomrule
    \end{tabular}
    \caption{Comparison of accuracy between different resolution selection strategies in the multi-region case.}
    \label{tab:ablation_select}
\end{table}

\begin{table}[t]
  \centering

  % 减小列间距以适应页面宽度，可根据你的文档边距微调
  \setlength{\tabcolsep}{4pt} 
  \begin{tabular}{l cc cc c}
    \toprule
    \multirow{2}{*}{\textbf{Design}} 
    % \cmidrule 用于创建跨越部分列的水平线
    & \multicolumn{2}{c}{\textbf{FSP}} & \multicolumn{2}{c}{\textbf{FCP}} & \multirow{2}{*}{Acc}  \\
    \cmidrule(lr){2-3} \cmidrule(lr){4-5} 
    & Acc & IoU & Acc & IoU &  \\
    \midrule
    IoU Reward          & 81.8            & 39.3       & 63.5           & 47.4            & 72.7  \\
    \midrule
    - IoU Thresh        & 79.0            & 29.3       & 64.2           & 39.0            & 71.6   \\
    - IoU Reward        & 79.0            & 30.6       & 63.7           & 37.7            & 71.2  \\
    \bottomrule
  \end{tabular}
  \caption{Comparison between different reward designs in Stage-II on HR-Bench 8k.}
  \label{tab:ablation_reward_design}
\end{table}

\section{More Preliminary Analysis}

To more rigorously demonstrate DeepEyes's insufficient attention to cropped regions, we also conducted the following experiments.  All experiments in this section are conducted with the maximum number of visual tokens set to 1,024.
\noindent\textbf{GT test.} We isolate samples from benchmarks where regions predicted by DeepEyes poorly cover GT ($Overlap \le 0.2$). As shown in Table \ref{more_preliminary_gt}, Replacing the regions of these samples with GT only yields returns within 10\% (significant growth be expected if model attend to these regions). This provides a more rigorous proof that DeepEyes cannot fully utilize cropped regions.

\noindent \textbf{Noise test.} We select samples where DeepEyes fails when forced to answer directly (without cropping pattern) but succeeds with cropping. As whon in Table \ref{more_preliminary_noise}, when replacing the cropped region with noise, the model still answers most of these samples correctly. This demonstrates that the performance gain is attributed to the cropping pattern (structural cues) rather than the enhanced visual information within the crops.

\section{More Ablation Studies}

\noindent\textbf{Data Utilization Strategy.}
As shown in the Table \ref{tab:ablation_data}, we compare two data usage strategies. The first strategy, `Mix Data', involves training the model in a single stage by mixing our collected data with the Visual Probe data at a 1:1 ratio. The second strategy, which we term `Stage-II', is the two-stage approach adopted. In the first stage, the model is trained exclusively on our collected data. Subsequently, in the second stage, we train the model with a small amount of Visual Probe data, incorporating a grounding reward. The results demonstrate that our proposed two-stage strategy achieves superior performance.

\noindent\textbf{Resolution Selection Strategy.}
To demonstrate the effectiveness of our resolution selection strategy in the first stage, we design two simple baselines, `Hard' and `Random', and compare them against our method on the multi-target subset (FCP / Spatial) of the dataset. This subset is chosen because it is more challenging than the single-target subset (FSP / Attr), and thus better highlights the efficacy of the information difference mechanism in our first stage. In the `Hard' baseline, we consistently select the most heavily downsampled image (where $max(h,w)=224$). In the `Random' baseline, we randomly select one image from the pool of all images generated by downsampling the original image at various scales. As shown in the Table \ref{tab:ablation_select}, our method outperforms both of these simple baselines.

\noindent\textbf{Reward Design in Stage-II.}
We compare different reward designs in Stage-II on the HR-Bench 8k. As shown in the Table \ref{tab:ablation_reward_design}, the first row presents our design where an IoU reward is employed during the second-stage training (without the $L_{1}$ reward). In the second row, we remove the condition that the IoU reward is only applied when the overlap exceeds a certain threshold. The experimental results demonstrate that retaining the conditional IoU reward (i.e., applying it only when the overlap is above the threshold) leads to higher IoU and accuracy. This suggests that this condition effectively guides the model to perform more precise cropping. Furthermore, the models trained with an IoU reward consistently outperform those without it in terms of accuracy and IoU. This indicates that the performance gain from the Stage-II training is primarily attributed to the IoU reward itself, rather than simply the introduction of additional training data.

\section{Benchmarks and Metrics Details}
Our method is evaluated on three benchmarks. The first, \textbf{HR-Bench 8K}
with an average resolution of 7680, which consists
of two sub-tasks: Fine-grained Single-instance Perception (FSP) and Fine-grained Cross-instance Perception (FCP). The 8K images are cropped around
the objects in question to produce \textbf{HR-Bench 4K}.
The third, $\boldsymbol{V^{*}}$, with an average resolution of 2246x1582, features
sub-tasks in attribute recognition (Attr) and spatial reasoning (Spatial). We evaluate our model on three datasets: hr-8k, hr-4k, and vstar. For all three benchmarks, the evaluation metric is \textbf{accuracy} (Acc), defined as the number of questions answered correctly. Additionally, to assess the cropping precision of our model, we also employ the \textbf{Intersection over Union} (IoU). This is calculated as the IoU between the model's predicted cropping box and the GT box of the question-relevant region.

\begin{table}[h]
  \centering
  %\small % 在双栏格式下使用 \small 或 \footnotesize 节省空间
  \label{tab:training_time}
  \begin{tabular}{l l r}
    \toprule
    \multicolumn{2}{l}{\textbf{Method}} & \textbf{Time (A100-hours)} \\
    \midrule
    \multicolumn{2}{l}{DeepEyes} & 480.0 \\
    \midrule
    \multirow{4}{*}{Ours} & Stage-I & 320.0 \\
                          & Stage-II & 96.0 \\
                          & Offline Res. Selection & 2.5 \\
                          \cmidrule(l){2-3} % 在第2到3列下方画线
                          & \textbf{Total} & \textbf{418.5} \\
    \bottomrule
  \end{tabular}
  \caption{Comparison of Training Time.}
  \label{tab_traintime}
\end{table}

\begin{table}[t]
    \centering
    \begin{tabular}{lc}
    \toprule
     \textbf{Parameter}                                           & \textbf{Value}          \\
    \midrule
      train batch size                                      & 256           \\
      rollout num per sample                                & 16           \\
      ppo mini batch size                                   & 32           \\
      ppo micro batch size per gpu                          & 2           \\
      rollout log prob micro batch size per gpu             & 4           \\
      ref log prob micro batch size per gpu                 & 4           \\
      single response max tokens                            & 2048           \\
      max turns                                             & 5           \\
      kl loss coef                                          & 0.0           \\
      entropy coeff                                         & 0.0           \\
      nodes num                                             & 1           \\
      gpus num per node                                     & 8           \\
      learning rate in Stage-I                                         & 1e-6           \\
      learning rate in Stage-II                                        & 5e-7           \\
      overlap threshold ($\tau$)                                & 0.9           \\
      acc reward weight ($r_{acc}$)                        & 0.8           \\
      format reward weight ($r_{format}$)                  & 0.2           \\
      tool call weight ($r_{tool}$)                        & 1.2           \\
    \bottomrule
    \end{tabular}
    \caption{The hyperparameters we used in the training pipeline.}
    \label{tab:hyper}
\end{table}

\section{Training Details}
We show the related hyper-parameters we use in Table \ref{tab:hyper}. We also compare our training time with that of DeepEyes as shown in Table \ref{tab_traintime}. Our total training time is shorter than that of DeepEyes. This is because we reduce the number of input visual tokens in the first stage by compressing the original images, accelerating the model rollout process.

\section{Visualization Analysis}

\begin{figure*}[htbp]
\centering
\includegraphics[width=\linewidth]{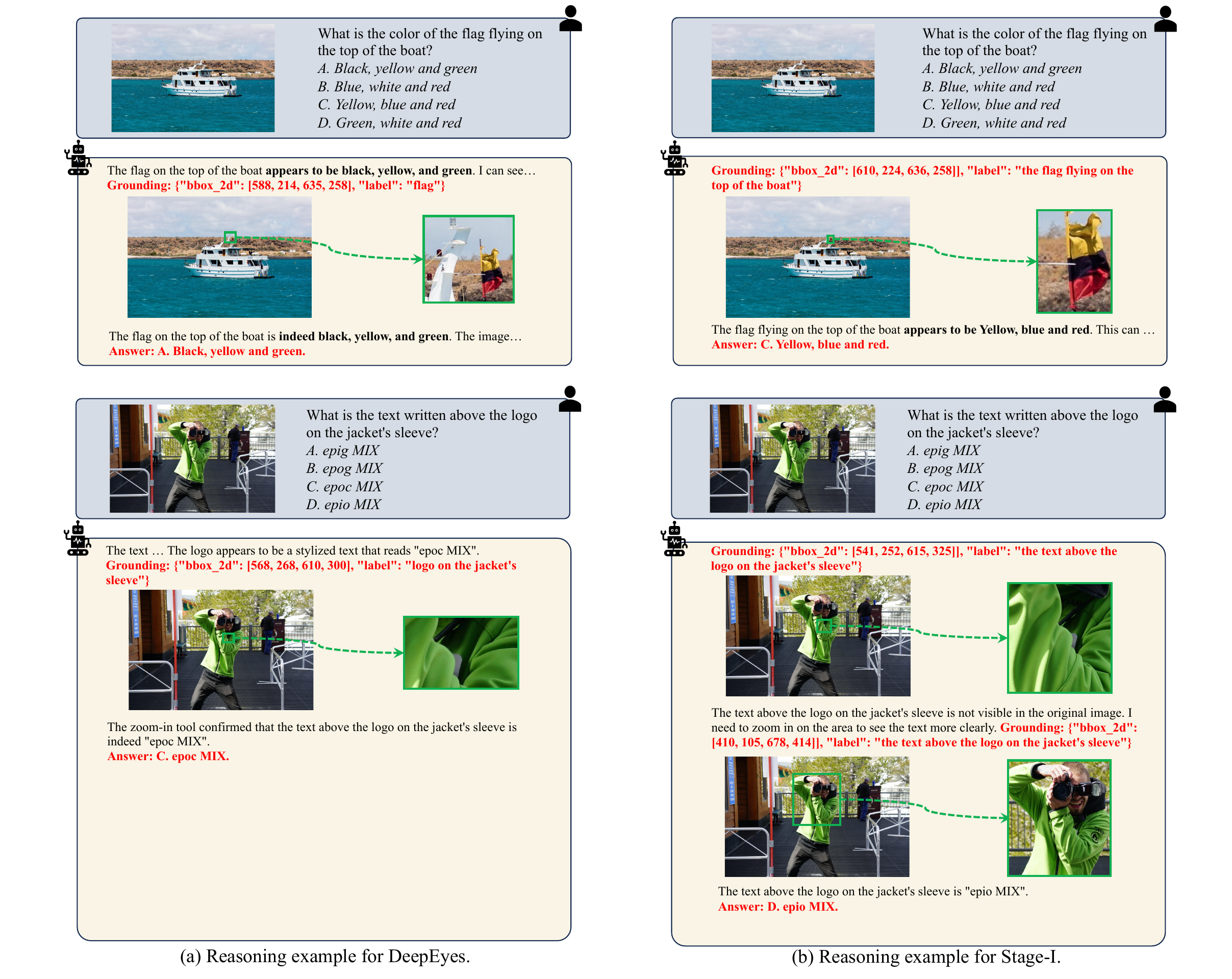} 
\caption{Comparison between DeepEyes and Stage-I.} 
\label{fig:vis_1}
\end{figure*}

\begin{figure*}
\centering
\includegraphics[width=\linewidth]{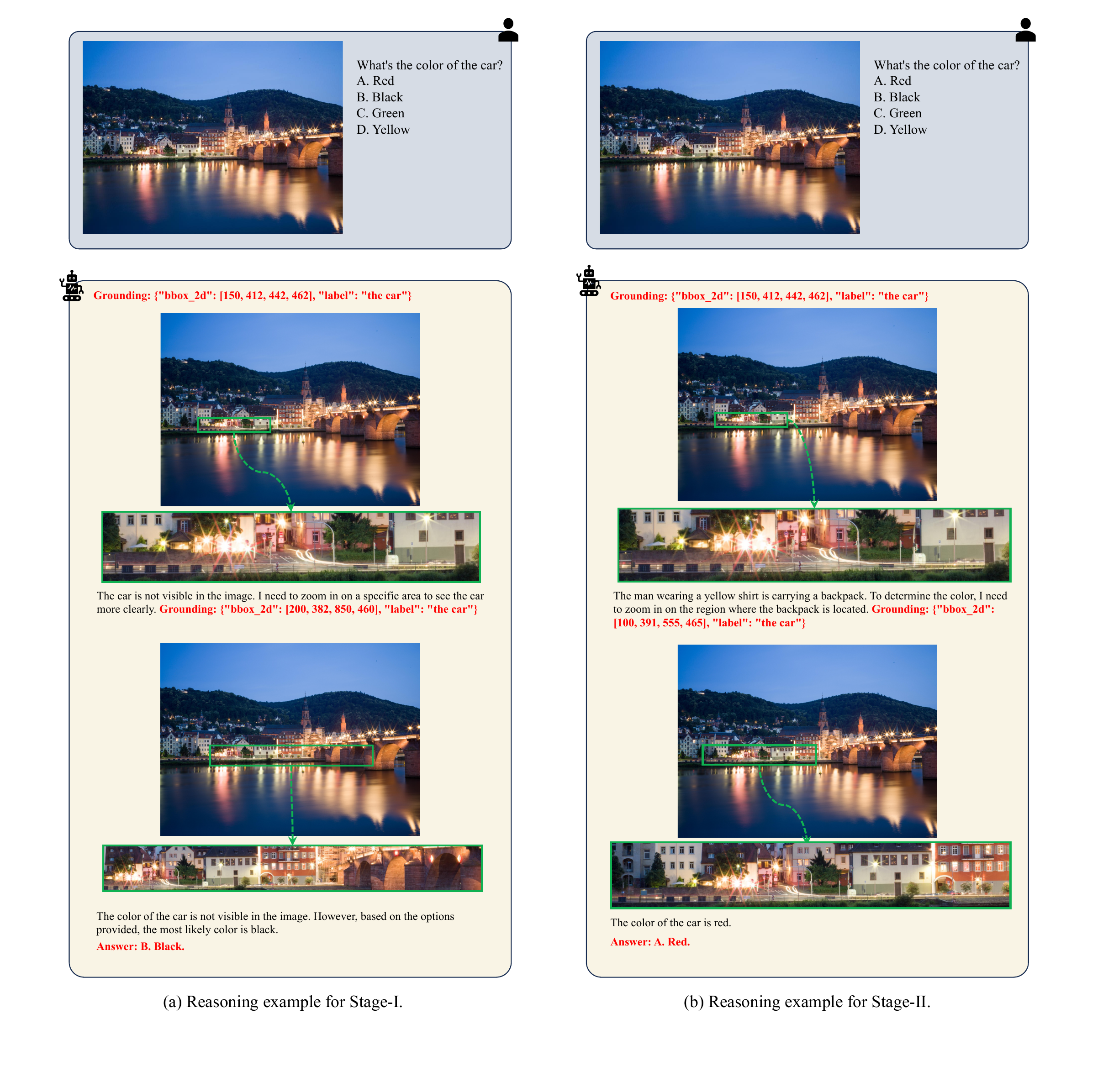} 
\caption{Comparison between Stage-I and Stage-II.} 
\label{fig:vis_2}
\end{figure*}

In Figure ~\ref {fig:vis_1}, we visually analyze the inference processes of DeepEyes and Stage-I. In the first example, while both models correctly crop the flag, DeepEyes provides an incorrect answer, whereas Stage-I arrives at the correct one. In the second example, both models initially fail to crop the jack's sleeve. However, DeepEyes proceeds to answer the question even with the jack's sleeve absent from the cropped region. In contrast, Stage-I identifies the absence of the jack's sleeve in the initial crop, performs a second cropping action, and ultimately succeeds in locating it and answering the question correctly. In Figure ~\ref {fig:vis_2}, we visualize and analyze the inference processes of the Stage-I and Stage-II models. In the first crop, both models fail to capture the target car and subsequently enlarge their cropping regions in the second attempt. However, the Stage-I model's second crop includes excessive redundant areas, causing it to fail again in identifying the car. In contrast, the Stage-II model accurately crops the entire road area, leading to a correct answer.

\section{Limitations and Future Works}

In this work, we first identify a critical issue in existing agent-based workflows for complex image understanding: the sub-optimal tool invocation that stems from a rigid formalization of the cropping tool. We address this by proposing an information gap mechanism. Building upon this, we further enhance model's cropping precision and, consequently, its overall performance by manually annotating a small set of bboxes and introducing a grounding reward. 

However, the performance gain from the second stage of training is relatively limited, which we attribute to the small number of annotated bboxes. This suggests that leveraging synthetic data \cite{wu2023datasetdm, zhao2023x}, could address the data scarcity problem in high-resolution VQA. For instance, a large-scale, high-resolution VQA dataset with bbox annotations could be synthesized by cropping objects from general-purpose VQA datasets like GQA \cite{hudson2019gqa} (which are low-resolution but rich in annotations including bbox) and pasting them onto high-resolution backgrounds. 

Furthermore, current methods only contains cropping tool, whose effectiveness is constrained by the inherent grounding ability of the MLLM and the fact that cropped regions contain background distractors. For future, a promising direction is to integrate segmentation models—such as open vocabulary semantic segmentation \cite{xu2023side, lan2024proxyclip, zhao2025training}, referring expression segmentation \cite{lai2024lisa, tang2025ufo}, or in-context segmentation \cite{wang2023seggpt, sheng2024towards, sheng2025unicl} —as callable tools for the agent.


\end{document}